\documentclass{article}

\usepackage[nonatbib,preprint]{neurips_2022}  

\usepackage[utf8]{inputenc} 
\usepackage[T1]{fontenc}    
\usepackage{url}            
\usepackage{booktabs}       
\usepackage{microtype}      
\usepackage{xcolor}         

\usepackage{amsmath,mathtools,amssymb,bbold}  
\usepackage{nicefrac}  
\usepackage{siunitx}  

\usepackage{graphicx}  
\usepackage[style=numeric-comp,sorting=none,backend=biber]{biblatex}  
\usepackage[bookmarks]{hyperref}  
\usepackage[nameinlink]{cleveref}  

\usepackage{subcaption}  

\usepackage{standalone}
\usepackage{tikz}
\usepackage[]{algorithm2e}

\SetKw{KwFrom}{from}
\SetKw{KwIn}{in}

\title{Sparse Kernel Gaussian Processes through Iterative Charted Refinement (ICR)}

\makeatletter
\newcommand{\printfnsymbol}[1]{%
	\textsuperscript{\@fnsymbol{#1}}%
}
\makeatother
\author{%
	Gordian Edenhofer\thanks{Gordian Edenhofer developed the idea, conducted the experiments and wrote parts of the manuscript. Reimar Leike had the initial idea and wrote parts of the manuscript.},\quad Reimar H. Leike\printfnsymbol{1}\thanks{This work was performed prior to employment by Amazon.},\quad Philipp Frank,\quad Torsten A. Enßlin\\
	Max Planck Institute for Astrophysics\\
	Karl-Schwarzschild-Straße 1\\
	85748 Garching bei München\\
	\texttt{\{edh,reimar,pfrank,ensslin\}@mpa-garching.mpg.de} \\
}

\begin{document}

\maketitle

\begin{abstract}
	Gaussian Processes (GPs) are highly expressive, probabilistic models.
	A major limitation is their computational complexity.
	Naively, exact GP inference requires $\mathcal{O}(N^3)$ computations with $N$ denoting the number of modeled points.
	Current approaches to overcome this limitation either rely on sparse, structured or stochastic representations of data or kernel respectively and usually involve nested optimizations to evaluate a GP.
	We present a new, generative method named Iterative Charted Refinement (ICR) to model GPs on nearly arbitrarily spaced points in $\mathcal{O}(N)$ time for decaying kernels without nested optimizations.
	ICR represents long- as well as short-range correlations by combining views of the modeled locations at varying resolutions with a user-provided coordinate chart.
	In our experiment with points whose spacings vary over two orders of magnitude, ICR's accuracy is comparable to state-of-the-art GP methods.
	ICR outperforms existing methods in terms of computational speed by one order of magnitude on the CPU and GPU and has already been successfully applied to model a GP with $122$ billion parameters.
\end{abstract}

\section{Introduction}
\label{sec:introduction}

Gaussian Processes (GPs) are flexible function approximators.
Their capacity to learn rich statistical representations scales with the amount of data they are provided with.
The statistical structure in the data is learned via the kernel.
It relates any two points in data-space and can be used to inter- and extrapolate.
Through their kernel, GPs are capable of representing intricate structures in large datasets.
One advantage of GPs is that they yield reliable uncertainty estimates when provided an adequate kernel.

The applicability of GPs however is limited by their scaling.
Naively evaluating a GP requires $\mathcal{O}(N^3)$ computations where $N$ is the number of modeled points.
The classical approach to GP inference requires applying the inverse of the kernel matrix (matrix representation of the kernel) and computing its log-determinant.
A common choice is to use Krylov subspace methods for both computations.
These methods are usually terminated after relatively few iterations, well before their theoretically guaranteed convergence.
In practice this reduces the computational complexity of evaluating a GP to $\mathcal{O}(n_\text{Kry} N^2)$ with $n_\text{Kry}$ the number of iterations of the Krylov subspace method. 

Several approaches exist to reduce the quadratic computational complexity for applying the kernel matrix.
They either require sparse kernels, structured kernels, regularly spaced data-points (modeled points), a set of inducing points, or a mixture of these.
Inducing point methods~\cite{WhenGaussianProcessesMeetsBigData:Liu2019} are popular because they do not require a special structure in the data or kernel.
The number of inducing points $M$ defines the method's ability to resolve structures.
It is desirable to choose $M \propto N$ which, however, renders their application impractical for large datasets with many modeled points.
Structured Kernel Interpolation (SKI) by \citeauthor{KernelInterpolation:Wilson2015} combines the advantages of regular grids and inducing points.
In \citetitle{KernelInterpolation:Wilson2015}~\cite{KernelInterpolation:Wilson2015} the authors use SKI to achieve $\mathcal{O}(N \log{N})$ computational complexity for applying the kernel matrix with $M=N$ inducing points.  

The application of the inverse as well as computation of the log-determinant of the kernel matrix remains costly within all these approaches.
Our goal is to reformulate modeling of GPs as generative process to reduce the computational complexity.
Without loss of generality, we shift the complexity of applying the inverse and evaluating the log-determinant of the kernel matrix to applying the ``square-root'' of it.
Evaluating a GP requires applying the square-root of the kernel matrix exactly twice, once for the forward pass and once for backpropagating the gradient.

In this paper we devise an efficient algorithm with $\mathcal{O}(N)$ computational complexity to apply an approximate square-root of a kernel matrix of a decaying kernel.
Our algorithm is applicable to GPs with nearly arbitrarily spaced points but especially efficient along translationally or rotationally invariant axes.
We motivate our approach from a generative perspective on GPs that is free of any nested optimizations.
In particular, we provide:
\begin{itemize}
	\item A generative approach to GPs which avoids inverting and taking the log-determinant of the kernel matrix.
	\item Iterative Charted Refinement (ICR), a generative method with $\mathcal{O}(N)$ computational complexity that requires no nested optimization to evaluate a GP.
	\item Our open-source implementation (BSD 2-Clause license) which is available at \url{https://gitlab.mpcdf.mpg.de/ift/nifty} as part of the Numerical Information Field Theory package (GPLv3 license)~\cite{nifty2022mpcdf}. Our implementation of ICR is written in python and uses JAX~\cite{jax2018github} to just-in-time compile code for the CPU and GPU.
\end{itemize}

We begin by briefly reviewing the literature of GPs in \autoref{sec:related_work}.
Afterwards, in \autoref{sec:background} we delve into generative models and a specific choice of reference frame called standardization.
We formulate our algorithm in \autoref{sec:iterative_charted_refinement}.
First we start with a regular grid before generalizing our approach to nearly arbitrarily spaced points.
In \autoref{sec:experiments} we highlight the benefits as well as the limitations of our approach in instructive experiments, paying special attention to KISS-GP.
Finally, we conclude in \autoref{sec:conclusion}.

\section{Related work}
\label{sec:related_work}

There are a wide variety of approaches to modeling GPs among which are full~\cite{ExactGaussianProcessesOnAMillionDataPoints:Wang2019}, sparse, structured sparse~\cite{FastKernelLearningForMultidimensionalPatternExtrapolation:Wilson2014,KernelInterpolationForScalableStructuredGaussianProcessesKISSGP:Wilson2015,ThoughtsOnMassivelyScalableGaussianProcesses:Wilson2015,SparseGaussianProcessesUsingPseudoInputs:Snelson2005,GPyTorch:Gardner2018,ConstantTimePredictiveDistributionsForGaussianProcesses:Pleiss2018}, and stochastic~\cite{GaussianProcessesForBigData:Hensman2013,StochasticVariationalDeepKernelLearning:Wilson2016} approaches.
We refer to the review by \citeauthor{WhenGaussianProcessesMeetsBigData:Liu2019}~\cite{WhenGaussianProcessesMeetsBigData:Liu2019} for a comprehensive discussion.
In the following, we would like to highlight a few notable developments which focus on the non-stochastic application of GPs to big data without strong grid or kernel constraints.
\looseness=-1

Without additional constraints, GPs require at least $\mathcal{O}(n_\text{solve} N^2)$ computations as they invoke $n_\text{solve}$ matrix-vector-multiplications (MVMs) of the matrix representation of the kernel to apply its inverse.
The required application of the inverse kernel matrix for evaluating a GP can be done efficiently with Krylov subspace methods, in particular preconditioned conjugate gradient and Lanczos methods~\cite{GPyTorch:Gardner2018,ConstantTimePredictiveDistributionsForGaussianProcesses:Pleiss2018,FastMatrixSquareRootsWithApplicationsToGaussianProcessesandBayesianOptimization:Pleiss2020}.
Furthermore, the computation can be distributed with minimal communication overhead~\cite{ExactGaussianProcessesOnAMillionDataPoints:Wang2019}.
However, neither gain in efficiency resolves the quadratic computational scaling inherent from applying and representing the kernel matrix.

Inducing point methods bypass the quadratic scaling with the number of modeled points by approximating the true kernel matrix $K_{XX}$ with the kernel matrix of the inducing points $K_{UU}$.
Either much fewer inducing points can be used than modeled points or their spacing can be chosen such that applying $K_{UU}$ becomes very efficient.
With KISS-GP, \citeauthor{KernelInterpolation:Wilson2015} propose a method with regularly spaced inducing points and Toeplitz and Kronecker structures in $K_{UU}$~\cite{KernelInterpolation:Wilson2015}.
The kernel matrix of the inducing points is mapped to the kernel matrix of the modeled points via a sparse interpolation matrix $W$.
Their approximation of the kernel matrix reads:
\begin{equation}
	K_{XX} \approx W K_{U U} W^T \ .
\end{equation}
In general, this matrix is singular and only becomes invertable in combination with projections (e.g. via a preconditioner) or additive corrections (e.g. diagonal jitter).
For $M=N$ inducing points as suggested by the authors and Toeplitz structure in the kernel matrix of the inducing points, applying the KISS-GP kernel matrix requires $\mathcal{O}(N \log{N})$ computations.

\section{Background}
\label{sec:background}

\subsection{Gaussian Processes}
\label{sec:background:gaussian_processes}

GPs are infinite dimensional stochastic processes for which every finite marginal is a multivariate normal distribution.
A GP is completely determined by its mean and kernel.
In the following we use $\mathcal{G}\left(\mu(x), k(x, x')\right)$ to refer to a GP with mean $\mu(x)$ and kernel $k(x, x')$ with $x$ respectively $x'$ denoting a location.
The amplitude of fluctuations and overall smoothness is determined by the kernel.

Popular kernels include the Radial Basis Function (RBF) and the Matérn kernel.
These kernels have various parameters that are to be inferred during optimization.
We denote these parameters by $\theta$ and the random vector drawn from $\mathcal{G}\left(\mu(x), k(x, x')\right)$ by $s$.
For any finite set of locations $X$ we denote the covariance kernel by $K_{XX}$ and the corresponding multivariate normal distribution by $\mathcal{N}(\mu(X), K_{XX})$.
The kernel matrix $K_{XX}$ depends on the parameters of the kernel $\theta$.
For brevity of notation, we do not make the dependence explicit.
Without loss of generality we assume the mean of our GP to be zero, i.e. $\mu(x) = 0$.

\subsection{Standardization}
\label{sec:background:standardization}

Inference with a GP is often formulated in terms of a Gaussian Process prior and a Gaussian likelihood with data $y$ and some, often diagonal, noise covariance.
In the following we focus on the more general problem with an arbitrary likelihood:
\begin{equation}
	\log{p(y,{(s,\theta)}^T)} = \log{p(y|s)} -\frac{1}{2} \left[ \log\left|2\pi K_{X X}\right| + s^T K_{X X}^{-1} s \right] + \log{p(\theta)} \label{eqn:general_joint_log_prob}\ .
\end{equation}
The quantity of interest is the posterior $p({(s,\theta)}^T|y)$.
In contrast to the case with Gaussian likelihood, no closed form solution exists even for a given $\theta$.
To perform an inference, the posterior is approximated.
A popular choice is to use Variational Inference (VI)~\cite{ADVI:Warren2017,MGVI:Knollmueller2019,geoVI:Frank2021}.

Naively, to evaluate \autoref{eqn:general_joint_log_prob} it is necessary to apply the inverse of the kernel as well as compute its log-determinant.
By a change of variables, these computations can be avoided.
In a generative approach with standardized parameters, $s$ and $\theta$ are expressed in terms of ``simpler'' random variables $\xi_s$ and $\xi_\theta$.
A common choice for the distribution of $\xi_{(\cdot)}$ is the standard Gaussian distribution.
The complexity of $s$ and $\theta$ is encoded into the deterministic mapping ${(s(\xi),\theta(\xi))}^T$ with $\xi \coloneqq (\xi_s, \xi_\theta)$.
This so-called standardization of ${(s,\theta)}^T$ is possible under very mild regularity conditions.
We refer to \citeauthor{VariationalInferenceWithNormalizingFlows:Rezende2015}~\cite{VariationalInferenceWithNormalizingFlows:Rezende2015} for a more detailed discussion on this subject.

Without loss of generality, we can rewrite equation \autoref{eqn:general_joint_log_prob} in terms of $\xi$:
\begin{equation}
	\log{p(y,\xi)} = \log{p(y|s(\xi))} -\frac{1}{2} \left[ \log\left|2\pi \mathbb{1}\right| + \xi^T \xi \right] \label{eqn:standardized_joint_log_prob}\ .
\end{equation}
Both inference problems are equivalent but note \autoref{eqn:standardized_joint_log_prob} involves no inversion nor a log-determinant of the kernel if $s(\xi)$ involves none.
Additionally, the standardization of the distribution for the parameters is numerically advantageous.
This is because after their standardization, all parameters are a priori defined on similar scales.

The expression for $s(\xi)$ implicitly computes the kernel parameters $\theta$.
Loosely speaking, we need the square-root of the kernel matrix $\sqrt{K_{XX}}$ to correlate the standard Gaussian random variables $\xi_s$: $s(\xi) = \sqrt{K_{XX}} \cdot \xi_s$.
The term $\sqrt{K_{xx}}$ denotes an implicit linear operator that correlates the values of $\xi_s$ such that ${\langle \left(\sqrt{K_{XX}} \cdot \xi_s\right) \left(\sqrt{K_{XX}} \cdot \xi_s\right)^{T} \rangle}_{p(\xi_s)} = \sqrt{K_{XX}} {\langle\xi\xi^T\rangle}_{p(\xi_s)} \sqrt{K_{XX}}^{T} = K_{XX}$ with $p(\xi_s)$ a standard normal distribution over $\xi_s$.
The kernel parameters are mapped via inverse transform sampling: $\theta(\xi_\theta) = \mathrm{CDF}^{-1}_{\theta}(\mathrm{CDF}_{\xi_\theta}(\xi_\theta))$ where $\mathrm{CDF_{(\cdot)}}$ denotes the cumulative density function of $p(\cdot)$.

The square-root of the kernel is not uniquely defined.
Any (potentially non-square) matrix $\sqrt{K_{XX}}$ which fulfills $\sqrt{K_{XX}} \sqrt{K_{XX}}^T = K_{x x}$ would suffice.
For arbitrary matrices applying some form of square-root of the matrix, e.g. via the Cholesky decomposition, is computationally as or even more expensive than applying its inverse.
In the following we devise an algorithm with $\mathcal{O}(N)$ complexity to efficiently apply an approximate $\sqrt{K_{XX}}$ for smoothly varying, decaying kernels.

\section{Iterative Charted Refinement}
\label{sec:iterative_charted_refinement}

The goal of this section is to outline an algorithm which applies an approximate version of $\sqrt{K_{XX}}$, which we denote by $\sqrt{K_\text{ICR}}$, to standard Gaussian distributed variable $\xi_s$ such that ${\langle \left(\sqrt{K_\text{ICR}}(\xi_s)\right)^{} \left(\sqrt{K_\text{ICR}}(\xi_s)\right)^{T} \rangle}_{p(\xi_s)} \approx K_{XX}$.
We start by deriving the method on a one dimensional, regular grid and later generalize it to points located in an arbitrarily spaced coordinate system.
Special attention is paid to how symmetries in the location of the modeled points can be utilized for additional computational performance improvements.

The core idea of our algorithm is to model the desired GP at varying resolutions simultaneously.
Each view of the GP at a given resolution is also a view on the correlations at a different length scale.
In a step of our iterative algorithm, the previous level is refined to a higher resolution until the maximum iteration depth is reached.

\subsection{Refining a three pixel grid}
\label{sec:icr:refining_a_three_pixel_grid}

\begin{figure*}[ht]
    \centering
	\begin{tikzpicture}
	[scale=0.75,transform shape,]

	\draw[thick] (0,0) -- (6,0);
	\draw[thick] (0,2) -- (6,2);

	\foreach \x in {0,...,3}
		{\draw[thick] (2*\x,0) -- (2*\x,2);}

	\draw[->, thick] (6.5,1) -- (9.5,1);
	\begin{scope}[shift={(10,0)}]
			\draw[thick] (0,0) -- (6,0);
			\draw[thick] (0,2) -- (6,2);

		\foreach \x in {0,...,3}
			{\draw[thick] (2*\x,0) -- (2*\x,2);}

			\draw[red, thick] (2,0) -- (4,0);
			\draw[red, thick] (2,2) -- (4,2);

		\foreach \x in {0,...,2}
			{\draw[red, thick] (\x+2,0) -- (\x+2,2);}
	\end{scope}
	\begin{scope}[shift={(0,-2.)}]
		\foreach \x in {0,...,2}
		{
			\draw[black, fill=black] (1+2*\x,1) circle (.1);
				\node at (1+2*\x,.5) {$x^\text{c}_\x$};
		}

		\draw[->, thick] (6.5,1) -- (9.5,1);
		\begin{scope}[shift={(10,0)}]
			\foreach \x in {0,...,2}
			{
				\draw[black, fill=black] (1+2*\x,1) circle (.1);
					\node at (1+2*\x,.5) {$x^\text{c}_\x$};
			}

			\draw[red, fill=red] (2.5,1.) circle (.1);
				\node at (2.5,.5) {$x^\text{f}_0$};
			\draw[red, fill=red] (3.5,1.) circle (.1);
				\node at (3.5,.5) {$x^\text{f}_1$};
		\end{scope}
	\end{scope}
\end{tikzpicture}
	\caption[1D Grid refinement]{
	Refinement of a $3$-pixel lattice.
	The upper row shows the grid refinement in terms of pixels, the lower row shows the grid refinement in terms of pixel centers.
	For GPs defined on these coordinates, one can analytically calculate the conditional distribution of the fine grid (red dots) given the coarse grid (black dots).
	\label{fig:refine_3_pixel_lattice}%
	}
\end{figure*}
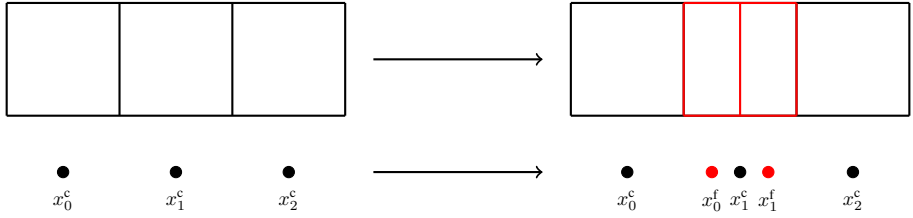

To derive the method, we constrain ourselves to a $3$ pixel sub-grid of a regular grid.
We would like to refine the central pixel of this sub-grid, as depicted in the upper row of \autoref{fig:refine_3_pixel_lattice}.
Hereby we interpret the pixel values as realizations of a GP at their respective centers (lower row of \autoref{fig:refine_3_pixel_lattice}).
The central property of a GP states that the $3+2$ coarse and fine pixels are jointly multivariate normal distributed:
\begin{align}
    \begin{pmatrix}
        s^\text{f}\\
        s^\text{c}
    \end{pmatrix}
    \sim\mathcal{N}\left(0,
    \begin{pmatrix}
        K_\text{ff} & K_\text{fc}\\
        K_\text{cf} & K_\text{cc}
    \end{pmatrix}
    \right) \ .
\end{align}
where $s^\text{f}$ is the realization of the GP on the two fine grid coordinates $x^\text{f}_i$ and $s^\text{c}$ is the realization of the GP at the three coarse grid positions $x^\text{c}_j$.
The entries of the covariance are given by
\begin{equation}
	\begin{aligned}
		K_\text{ff} &=
		\begin{pmatrix}
			k(x^\text{f}_0, x^\text{f}_0) & k(x^\text{f}_0, x^\text{f}_1)\\
			k(x^\text{f}_1, x^\text{f}_0) & k(x^\text{f}_1, x^\text{f}_1)
		\end{pmatrix} \ ,\\
		K_\text{fc} = K_\text{cf}^T &=
		\begin{pmatrix}
			k(x^\text{f}_0, x^\text{c}_0) & k(x^\text{f}_0, x^\text{c}_1) & k(x^\text{f}_0, x^\text{c}_2)\\
			k(x^\text{f}_1, x^\text{c}_0) & k(x^\text{f}_1, x^\text{c}_1) & k(x^\text{f}_1, x^\text{c}_2)
		\end{pmatrix} \ ,\\
		K_\text{cc} &=
		\begin{pmatrix}
			k(x^\text{c}_0, x^\text{c}_0) & k(x^\text{c}_0, x^\text{c}_1) & k(x^\text{c}_0, x^\text{c}_2)\\
			k(x^\text{c}_1, x^\text{c}_0) & k(x^\text{c}_1, x^\text{c}_1) & k(x^\text{c}_1, x^\text{c}_2)\\
			k(x^\text{c}_2, x^\text{c}_0) & k(x^\text{c}_2, x^\text{c}_1) & k(x^\text{c}_2, x^\text{c}_2)
		\end{pmatrix}
		\label{eqn:refinement_kernel_coarse_fine_combinations}
		\ .
	\end{aligned}
\end{equation}

Given the realization of the GP at the coarse grid locations $x^\text{c}_i$, the conditional distribution of the fine grid points $x^\text{f}_j$ is given by
\begin{align}
    p(s^\text{f}|s^\text{c}) &= \mathcal{N}(s^\text{f}|Rs^\text{c}, D)\, ,\\
	\text{with } R &= K_\text{fc} K^{-1}_\text{cc} \label{eqn:refinement_filter} \\
	\text{and } D &= K_\text{ff} - K_\text{fc} K^{-1}_\text{cc} K_\text{cf} \label{eqn:refinement_propagator}\ .
\end{align}
One can reformulate this into a generative process of the fine grid realizations $s^\text{f}$ given the coarse grid realizations $s^\text{c}$:
\begin{align}
    s^\text{f} = Rs^\text{c} + \sqrt{D}\xi_s^f \label{eq:generative-formulation-small}
\end{align}
where $\sqrt{D}$ is the Cholesky decomposition of $D$ and $\xi_s^f$ is a two-dimensional independent standard normal distributed vector.
We call $R$ and $\sqrt{D}$ the refinement matrices.
The generative formulation for $s^\text{f}$ using the refinement matrices is linear in $(s^\text{c}, \xi_s^f)$.

\subsection{Larger lattices}
\label{sec:icr:larger_lattices}

So far we have only considered refining the center pixel of a three pixel grid.
In this subsection we extend this notion to arbitrary large grids.
The main idea is to refine every pixel of the coarse grid using the procedure outlined in \autoref{sec:icr:refining_a_three_pixel_grid}, except for the outermost two pixels, as they do not have sufficient neighbors to be refined in that way.
Since two fine pixels emerge from refining one coarse pixel, the number of fine pixels $N^\text{f}$ is $2 \cdot (N^\text{c} - 2)$, where $N^\text{c}$ is the number of coarse pixels.

This way of refining the grid is an approximation, as ideally every fine pixel would be informed by every coarse pixel, not just by its $3$ nearest coarse pixels.
The approximation can be understood in terms of probability distributions:
\begin{gather}
	p((s^\text{f}_i)_{i=0\dots N^\text{f}-1}|(s^\text{c}_i)_{i=0\dots\nicefrac{N^\text{f}}{2}+1}) \approx p_\text{approx}(s^\text{f}|s^\text{c}) \quad\text{with}\nonumber\\
	p_\text{approx} = p((s^\text{f}_i)_{i=0,1}|(s^\text{c}_i)_{i=0,1,2})\times \dots \times p((s^\text{f}_i)_{i=N^\text{f}-1,N^\text{f}}|(s^\text{c}_i)_{i=\nicefrac{N^\text{f}}{2}-1,\nicefrac{N^\text{f}}{2}, \nicefrac{N^\text{f}}{2}+1}) \label{eqn:refinement_core_approximation}\ .
\end{gather}
This approximation looses information in two different ways.
On the one hand, factoring $P((s^\text{f}_i)_{i=0\dots N^\text{f}}|(s^\text{c}_i)_{i=0\dots\nicefrac{N^\text{f}}{2}+1})$ with respect to pairs of points $(s^\text{f}_i)_{i=2j,2j+1}$ neglects correlations of the fine grid given the coarse grid.
Furthermore, information about the fine grid points' dependence on coarse grid points at larger distances is neglected.
Both approximations are expected to be less severe if there are no systematic long range correlations when conditioning on the coarse grid.

This approximative way to calculate the fine grid points has computational advantages.
It enables calculating all pairs of fine grid points independently in a similar way.
On a regular grid, if the kernel $k$ is stationary, i.e. $k(x,y)=k(x-y)$, then all probability distributions of the product in \autoref{eqn:refinement_core_approximation} are the same, only evaluated for different variables.
We can formulate the refinement once again as a generative process, by following the procedure of \autoref{eq:generative-formulation-small} for every pair of fine grid points.
Doing so can be interpreted as a convolution, with $R_{oj}$ being the convolution kernel of size $2 \times 3$, such that
\begin{gather}
    r_{2i+o} = \sum_{j=0}^2R_{oj} s^\text{c}_{i+j} \label{eqn:generative_refinement_conv}
    \\ s^\text{f}_{2i+o} = r_{2i+o} + \sum_{j=0}^1\sqrt{D}_{oj}(\xi_s)_{ij} \label{eqn:generative_refinement_correction}
\end{gather}
with $r$ being the filtered coarse grid, $o\in\{0,1\}$ and $(\xi_s)_{ij}$ an independent standard normal distributed vector.
\autoref{eqn:generative_refinement_conv} and \autoref{eqn:generative_refinement_correction} show similarities to multigrid approaches, which are used for simulations~\cite{ScientificComputing:Heath2018,AnIntroductionToMultigridMethods:Wesseling2004,OnTheConvergenceOfARelaxationMethodWithNaturalConstraintsOnTheEllipticOperator:Bakhvalov1966}.
The values of the fine grid are constructed through smoothing the values of the coarse grid with the kernel $R$ \eqref{eqn:generative_refinement_conv} and then applying a correction \eqref{eqn:generative_refinement_correction}.

The approximate probability distribution $p_\text{approx}(s^\text{f}|s^\text{c})$ allows us to draw a sample of $s^\text{f}$ given the realization $s^\text{c}$ on the coarse grid.
One can iterate this procedure, using the fine lattice as a coarse lattice to an even finer lattice.
This gives rise to a generative process, depicted in \autoref{fig:refinement_layers_regular}.
Hereby the total grid size shrinks by 2 pixels in every refinement, but all other pixels are split in two.

The overall cost of generating a sample when performing iterative grid refinement $n_\text{lvl}$ times on an initial grid with $N^{(0)}$ grid points is
\begin{align}
	3N^{(0)} + 6(N^{(0)}-2) + 6(2(N^{(0)}-2) \cdot 2 - 2) + \underbrace{\dots}_\text{$n_\text{lvl} - 2$ additional terms} \ .
\end{align}
For $N^{(0)}>4$ this is in $\mathcal{O}(2^{n_\text{lvl}} \cdot N^{(0)})=\mathcal{O}(N)$, where $N$ is the number of pixels of the final lattice.
One can start this process from an arbitrarily coarse grid with at least $3$ pixels for which the covariance matrix can be diagonalized explicitly at negligible computational cost.
The overall grid refinement algorithm is summarized in \autoref{alg:grid_refinement}.
It can generate approximate samples from a GP with $\mathcal{O}(N)$ computations where $N$ is the number of pixels of the final lattice.

\begin{algorithm}[ht]
	\KwData{$\xi_s^{(0)}\dots\xi_s^{(n)}$, $\sqrt{D}^{(0)}\dots\sqrt{D}^{(n)}$, $R^{(1)}\dots R^{(n-1)}$}
	\KwResult{$s^n$: A sample with approximate correlation structure $K_{XX}$}
	$s^{(0)}_i \leftarrow \sum_j \sqrt{D^{(0)}_{ij}} (\xi_s^{(0)})_j$\;
	$N^{(0)} \leftarrow \text{size}(s^{(0)})$\;
	\For{$l$ \KwFrom $1$ \KwTo $n_\text{lvl}$}{
		$N^{(l)} \leftarrow 2(N^{(l-1)}-2)$\;
		\For{$i$ \KwFrom $0$ \KwTo $N^{(l-1)}-2$}{
			\For{$o$ \KwIn $\{0,1\}$}{
				$r^{(l)}_{2i+o} \leftarrow \sum_{j=0}^2R_{oj} s^{(l-1)}_{i+j}$\;
				$s^{(l)}_{2i+o} \leftarrow r^{(l)}_{2i+o} + \sum_{j=0}^{1}\sqrt{D}_{oj}(\xi_s^{(l)})_{ij}$\;
			}
		}
	}
	\caption{
	Iterative grid refinement of a one-dimensional regular grid with stationary kernel and open boundary conditions.
	\label{alg:grid_refinement}%
	}
\end{algorithm}

\subsection{Generalization to arbitrary grid locations}
\label{sec:icr:generalization_to_arbitrary_grid_locations}

To generalize the above treatment to arbitrarily spaced points, it is convenient to employ the concept of a coordinate chart from Topolgy.
A coordinate chart is a homeomorphism from an open subset $\mathcal{D}$ of a manifold, i.e. the space on which our GP is defined, to an open subset of a Euclidean space.
Every location in $\mathcal{D}$ is associated with a unique location in Euclidean space via a coordinate chart.

ICR requires the user to specify a coordinate chart together with the data.
For many modeling problems, such a chart is straightforward to formulate, e.g. a pixel detector measuring energies might have a regular, spatial pixel axis and a logarithmic, spectral energy axis.
However, for general data, defining such a chart requires expert knowledge and is left to the user.
In the following we use $\phi^{-1}$ to denote a coordinate chart from a Euclidean grid to $\mathcal{D}$.
The kernel acts on points in $\mathcal{D}$ while the refinement is carried out on a Euclidean grid.

\newsavebox{\largestimage}
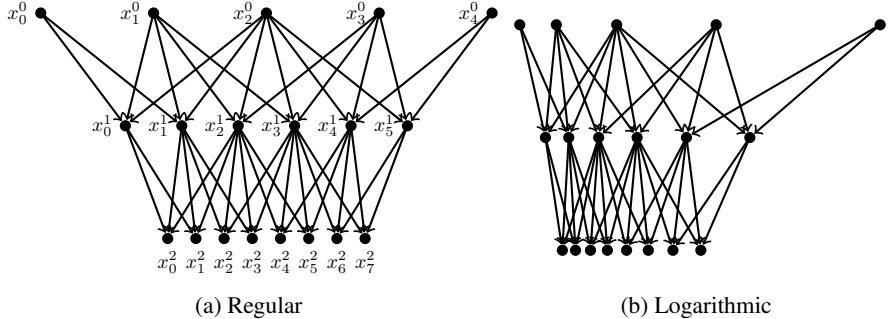
\begin{figure}[ht]
    \centering
	\savebox{\largestimage}{\begin{tikzpicture}
	[scale=0.75,transform shape,]

    \foreach \x in {0,...,4}
    {
        \node[label=left:{$x^0_\x$},circle,fill,inner sep=.07cm] (0\x) at (1+2*\x,1) {};
    }

    \foreach \x in {0,...,5}
    {
        \node[label=left:{$x^1_\x$},circle,fill,inner sep=.07cm] (1\x) at (2.5+\x,-1) {};
    }
    \foreach \x in {0,...,5}
    {
        \pgfmathparse{div(\x,2)}\let\tf\pgfmathresult
        \draw[thick, ->] (1+2*\tf, 1) -- (1\x);
        \draw[thick, ->] (3+2*\tf, 1) -- (1\x);
        \draw[thick, ->] (5+2*\tf, 1) -- (1\x);
    }

    \foreach \x in {0,...,7}
    {
        \node[label=below:{$x^2_\x$},circle,fill,inner sep=.07cm] (2\x) at (3.25+.5*\x,-3) {};
    }
    \foreach \x in {0,...,7}
    {
        \pgfmathparse{div(\x,2)}\let\tf\pgfmathresult
        \draw[thick, ->] (2.5+\tf, -1) -- (2\x);
        \draw[thick, ->] (3.5+\tf, -1) -- (2\x);
        \draw[thick, ->] (4.5+\tf, -1) -- (2\x);
    }
\end{tikzpicture}}
	\subcaptionbox{Regular\label{fig:refinement_layers_regular}}{\usebox{\largestimage}}
	\vspace{0em}
	\subcaptionbox{Logarithmic\label{fig:refinement_layers_irregular}}{%
		\raisebox{\dimexpr.5\ht\largestimage-.5\height}{%
\tikzset{declare function={Chart(\rg)=1 * (exp(0.5 * (\rg)) - 1.);}}

\begin{tikzpicture}
	[scale=0.75,transform shape,]

    \foreach \x in {0,...,4}
    {
		\pgfmathparse{Chart(\x)}\let\tf\pgfmathresult
        \node[circle,fill,inner sep=.07cm] (0\x) at (\tf,1) {};
    }

    \foreach \x in {0,...,5}
    {
		\pgfmathparse{Chart(\x / 2 + 0.75)}\let\tf\pgfmathresult
        \node[circle,fill,inner sep=.07cm] (1\x) at (\tf,-1) {};
    }
    \foreach \x in {0,...,5}
    {
		\pgfmathparse{Chart(div(\x, 2))}\let\tcf\pgfmathresult
		\pgfmathparse{Chart(div(\x, 2) + 1)}\let\tcs\pgfmathresult
		\pgfmathparse{Chart(div(\x, 2) + 2)}\let\tct\pgfmathresult
        \draw[thick, ->] (\tcf, 1) -- (1\x);
        \draw[thick, ->] (\tcs, 1) -- (1\x);
        \draw[thick, ->] (\tct, 1) -- (1\x);
    }

    \foreach \x in {0,...,7}
    {
		\pgfmathparse{Chart(\x / 4 + 0.75 + 0.75 / 2)}\let\tf\pgfmathresult
        \node[circle,fill,inner sep=.07cm] (2\x) at (\tf,-3) {};
    }
    \foreach \x in {0,...,7}
    {
		\pgfmathparse{Chart(div(\x, 2) / 2 + 0.75)}\let\tcf\pgfmathresult
		\pgfmathparse{Chart(div(\x, 2) / 2 + 1.25)}\let\tcs\pgfmathresult
		\pgfmathparse{Chart(div(\x, 2) / 2 + 1.75)}\let\tct\pgfmathresult
        \draw[thick, ->] (\tcf, -1) -- (2\x);
        \draw[thick, ->] (\tcs, -1) -- (2\x);
        \draw[thick, ->] (\tct, -1) -- (2\x);
    }
\end{tikzpicture}%
		}
	}
    \caption{%
	Multiple grid refinements of a larger lattice.
	The left figure depicts the pixel centers on a regular grid within two refinement steps.
	The fine grid realizations are calculated on the basis of the three nearest, coarse grid realizations.
	Arrows indicate which coarse grid values inform which fine grid values.
	On the right, the same refinement translated with a logarithmic coordinate chart is shown.
	\looseness=-1
	\label{fig:refinement_layers}%
	}
\end{figure}

Our method as outlined in \autoref{sec:icr:larger_lattices} is agnostic with respect to the choice of coordinate system (see \autoref{fig:refinement_layers_irregular}) but the refinement matrices $R$ and $D$ are not.
The refinement matrices encode distances between points and depend on the parametrization of the kernel.
For arbitrarily spaced points, every pixel needs its own set of refinement matrices.
Furthermore, if the parameters of the kernel change, the refinement matrices must be recomputed.

To extend our algorithm to charted coordinate systems, we need to adapt \autoref{eqn:refinement_kernel_coarse_fine_combinations}.
Namely, instead of plugging in equidistant points $x_{j}^f$ and $x_{i}^c$, we first need to map our regular coordinates to the coordinate system of the modeled points.
We can do so by updating our definition of the kernel in \autoref{eqn:refinement_kernel_coarse_fine_combinations}.
Our new kernel reads $\tilde{k} : (\mathbb{R}, \mathbb{R}) \rightarrow \mathbb{R}$ with $\tilde{k}: (\tilde{x}, \tilde{x}') \mapsto k(\phi^{-1}(\tilde{x}), \phi^{-1}(\tilde{x}'))$.
With this minimally invasive update, our discussion in \autoref{sec:icr:refining_a_three_pixel_grid} and \autoref{sec:icr:larger_lattices} adapts to nearly arbitrarily spaced points.

The computational costs associated with constructing the refinement matrices for all pixels scale with $\mathcal{O}(\left(3^d\right)^3 \cdot N)$ where $d$ denotes the number of dimensions.
The cubic scaling in the number of coarse pixels $3^d$ in a refinement step stems from the explicit inversion of the covariance of the coarse pixels in \autoref{eqn:refinement_filter} and \autoref{eqn:refinement_propagator}.
If the kernel factorizes along certain dimensions, the computational complexity can be significantly reduced.
Likewise, for rotationally or translationally invariant axes within $\mathcal{D}$ and stationary kernel, the refinement matrices need only be computed once and can be broadcasted along these axes akin to \autoref{eqn:generative_refinement_conv} and \autoref{eqn:generative_refinement_correction}.
For commonly used coordinate systems like polar or spherical coordinates, significant performance improvements can be obtained by utilizing these symmetries.

\subsection{Generalization to larger refinement matrices}
\label{sec:icr:generalization_to_larger_refinement_matrices}

How much information we neglect at a refinement level depends on how many adjacent pixels we use to construct the refinement matrices.
In \autoref{sec:icr:refining_a_three_pixel_grid} and \autoref{sec:icr:larger_lattices} we refined $3$ coarse pixels to $2$ fine pixels.
The number of coarse pixels, the number of fine pixels as well as the position of the fine pixels on our Euclidean grid can be tuned.
The optimal setting for these parameters depends on the kernel and the coordinate chart.
In our open-source implementation which is available at \url{https://gitlab.mpcdf.mpg.de/ift/nifty} as part of the Numerical Information Field Theory package~\cite{nifty2022mpcdf}, we provide helper utilities to retrieve the information theoretical optimal settings for a given kernel and chart.
With more coarse and fine grid cells in a refinement, the construction of the refinement matrices gets more expensive.
It requires $\mathcal{O}\left(\max{\{n_\text{csz},n_\text{fsz}\}}^{3d} \cdot N\right)$ computations where $n_\text{csz}$ denotes the number of coarse pixels to refine to $n_\text{fsz}$ fine pixels.

\section{Experiments}
\label{sec:experiments}

We evaluate the accuracy and speed of our algorithm in \autoref{sec:experiments:comparison_to_excat_GPs} and \autoref{sec:experiments:comparison_to_KISS_GP} respectively.
In particular, we compare our implicit representation of the covariance against the true realization for a low number of modeled points for which explicitly instantiating the kernel matrix is possible.
Furthermore, we compare against KISS-GP in terms of speed.
We choose KISS-GP because it has a similar computational complexity and promises to be applicable to millions or even billions of data-points (modeled points), outperformed previous popular inducing point methods in terms of speed and accuracy~\cite{KernelInterpolationForScalableStructuredGaussianProcessesKISSGP:Wilson2015,ThoughtsOnMassivelyScalableGaussianProcesses:Wilson2015}, and it highlights the conceptual differences in modeling GPs compared to our approach.
\looseness=-1


In the following we assume an irregular grid with logarithmically spaced points.
We choose a logarithmic axis to determine the accuracy of our refinement method at various length scales.
As kernel, we use the homogeneous and isotropic Matérn covariance with degree-of-freedom parameter $\nicefrac{3}{2}$:
\looseness=-1
\begin{equation}
	k(d) = \left. \left(1 + \frac{\sqrt{3}d}{\rho}\right) \cdot \exp{\left(-\frac{\sqrt{3}d}{\rho}\right)} \right|_{\rho=\rho_0}\label{eqn:experiment_matern_kernel} \ ,
\end{equation}
where $d$ denotes the distances between two points and $\rho$ the characteristic length scale.


\subsection{Comparison to exact Gaussian Processes}
\label{sec:experiments:comparison_to_excat_GPs}

By construction, our kernel defined by $\sqrt{K_\text{ICR}} \sqrt{K_\text{ICR}}^T$ yields a positive semidefinite kernel matrix.
Thus, even though we approximate $K_{XX}$, the resulting process in the limit of infinite refinement depth is still a GP.
For any discrete realization, we can quantify the information loss due to our approximation with the Kullback-Leibler divergence~\cite{FoundationsOfInference:Knuth2012,OptimalBeliefApproximation:Reimar2017}.
We use this measure to pick the optimal set of refinement parameters (see~\autoref{sec:icr:generalization_to_larger_refinement_matrices}) within $(n_\text{csz}, n_\text{fsz})\in\{(3, 2), (3, 4), (5, 2), (5, 4), (5, 6)\}$ with $N=200$ and $n_\text{lvl}=5$ for a GP on logarithmically spaced points and a Matérn covariance as specified in \autoref{eqn:experiment_matern_kernel}.
Our modeled points are spaced such that the distances to nearest neighbors range from $2\%$ of $\rho_0$ up to $\rho_0$.
The fine pixels within each parametrization of ICR are located around the coarse center pixel and spaced such that they take up half the volume on the Euclidean grid of a coarse pixel.
We find the optimal refinement parameters to be $n_\text{csz}=5$ and $n_\text{fsz}=4$.
\looseness=-1

The top row of \autoref{fig:covariance_comparison_icr_vs_kiss_gp_matern_kernel} compares the implicit covariance matrix of ICR to the true one for $200$ modeled points.
Both covariance matrices are overall in agreement.
The diagonal is approximated comparatively well with errors of up to $6.5\cdot10^{-2}$.
In a regular pattern, there are significant differences in the strength of correlations with absolute errors up to $0.13$ ($13\%$ of the variance).
The mean absolute error (MAE) is $5.8\cdot10^{-3}$.
The absolute pixel wise errors are of comparable magnitude irrespective of the spacing between points.

The observable mismatch is the sum of two main sources of errors.
The first source of error lies within the refinement step itself.
Within a refinement, we disregard correlations by interpolating with the refinement matrix $R$ which uses only $n_\text{csz}$ neighboring coarse pixels.
Furthermore, we add small corrections to the previous level but only correlate these with the refinement matrix $\sqrt{D}$ in blocks of $n_\text{fsz}$ fine pixels.
Both approximations strictly decrease the strength of correlations between pixels.

The second source of error lies in the iterative nature of our algorithm.
In each refinement step we assume the previous refinement level to have modeled our desired GP without error.
As outline above, this assumption does not hold after the first iteration.
The interpolation matrix $R$ in our refinement mixes values of which the variance may be overestimated and the correlation underestimated.
Already after the second refinement level, the incurred errors due to our approximations within the algorithm are smeared out and potentially amplified.

\subsection{Comparison to KISS-GP}
\label{sec:experiments:comparison_to_KISS_GP}

In contrast to our method, KISS-GP must not necessarily produce a proper GP prior because their approximate representation of $K_{XX}$ is not always full rank.
This is the case for $M < N$ but also for $M \geq N$ and points with vastly different spacing between modeled points such that the interpolation does not use at least $M - N + 1$ of the regularly spaced inducing points.
To apply the inverse of the kernel matrix, it is necessary to add some small diagonal correction or use an appropriate preconditioner respectively projection operator.

\begin{figure}[ht]
	\centering
	\includegraphics[trim={0.0cm .3cm 0.0cm 0.2cm},clip,width=\linewidth]{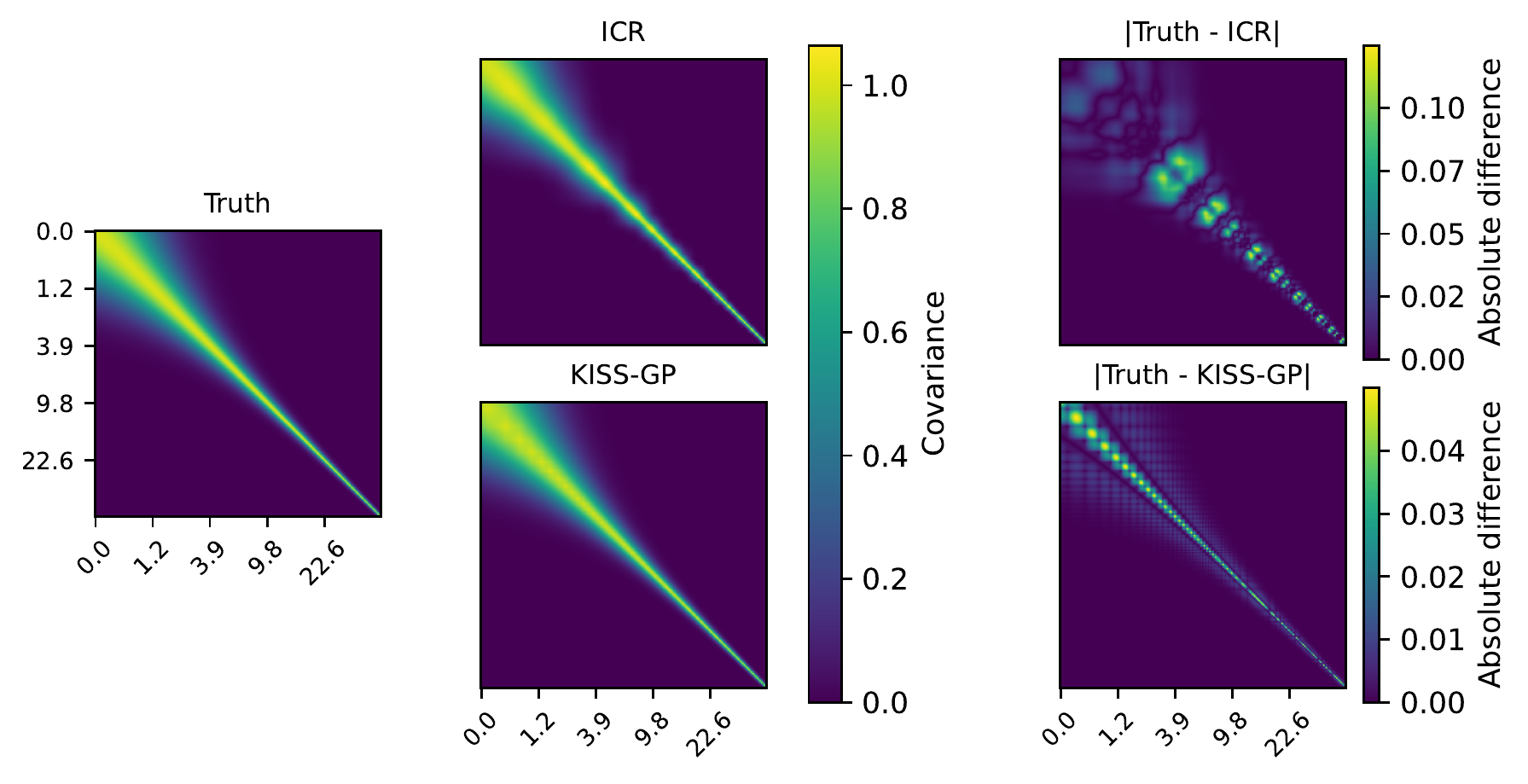}
	\caption{%
		The true covariance, the implicit covariance of ICR and KISS-GP as well as their respective absolute difference to the true one.
		Axes are logarithmic. Their labels are given in multiples of $\rho_0$.
		\label{fig:covariance_comparison_icr_vs_kiss_gp_matern_kernel}%
	}
\end{figure}

The bottom row of \autoref{fig:covariance_comparison_icr_vs_kiss_gp_matern_kernel} compares the covariance of KISS-GP to the true one for the same setting as in \autoref{sec:experiments:comparison_to_excat_GPs}.
We represent the KISS-GP covariance $K_\text{KISS-GP}$ in the harmonic domain with $M=N=200$ inducing points and a padding factor of $0.5$ to reduce the effects of periodic boundary conditions:
\begin{equation}
	K_\text{KISS-GP}=W \cdot \mathcal{F} \cdot P \cdot \mathcal{F}^T \cdot W^T \label{eqn:harmonic_ski}
\end{equation}
with $W$ denoting a sparse linear interpolation matrix, $\mathcal{F}$ the harmonic transformation and $P$ the harmonically transformed kernel from \autoref{eqn:experiment_matern_kernel}.
In terms of the absolute pixel wise difference in the covariance matrix for the above case, the approximation by KISS-GP is more accurate compared to our algorithm.
Its mean absolute error is $1.8\cdot10^{-3}$, which is $31\%$ of ICR's MAE.
The maximum absolute error is $4.9\cdot10^{-2}$ ($39\%$ of ICR's) and occurs on the diagonal ($75\%$ of ICR's maximum error on the diagonal).
The errors decrease if points are spaced more similarly to the evenly spaced inducing points, and significantly increase for spacings varying over several orders of magnitude.
\looseness=-1

Let us now compare the computational speed of ICR versus KISS-GP.
All experiments were carried out with double precision using $8$ cores of an Intel Xeon IceLake-SP 8360Y CPU with $62$ GB of RAM and a single NVIDIA A100 GPU with $40$ GB or HBM2.
We compare their respective performance on both the CPU and GPU.
For each method we time the execution of a single forward pass of the model.
In the case of our algorithm, we time the application of $\sqrt{K_\text{ICR}}$.
For KISS-GP a forward pass involves applying the square-root as well as computing the log-determinant of the kernel matrix.

We use relatively few Krylov subspace iterations to create a favorable setting for KISS-GP.
Irrespective of the number of modeled points, we use $40$ CG iterations to apply the inverse of the kernel matrix, and $10$ samples each optimized for $15$ Lanczos iterations to stochastically estimate the log-determinant.
We use $M=N$ inducing points and the KISS-GP approximation from \autoref{eqn:harmonic_ski} with no padding.
In practice, we would need to pad the domain of the inducing points and for a kernel in position space, we would also need at least one more harmonic transformation~\cite{ThoughtsOnMassivelyScalableGaussianProcesses:Wilson2015}.

\begin{figure}[ht]
	\centering
	\includegraphics[trim={0.0cm .2cm .0cm .0cm},clip,width=\linewidth]{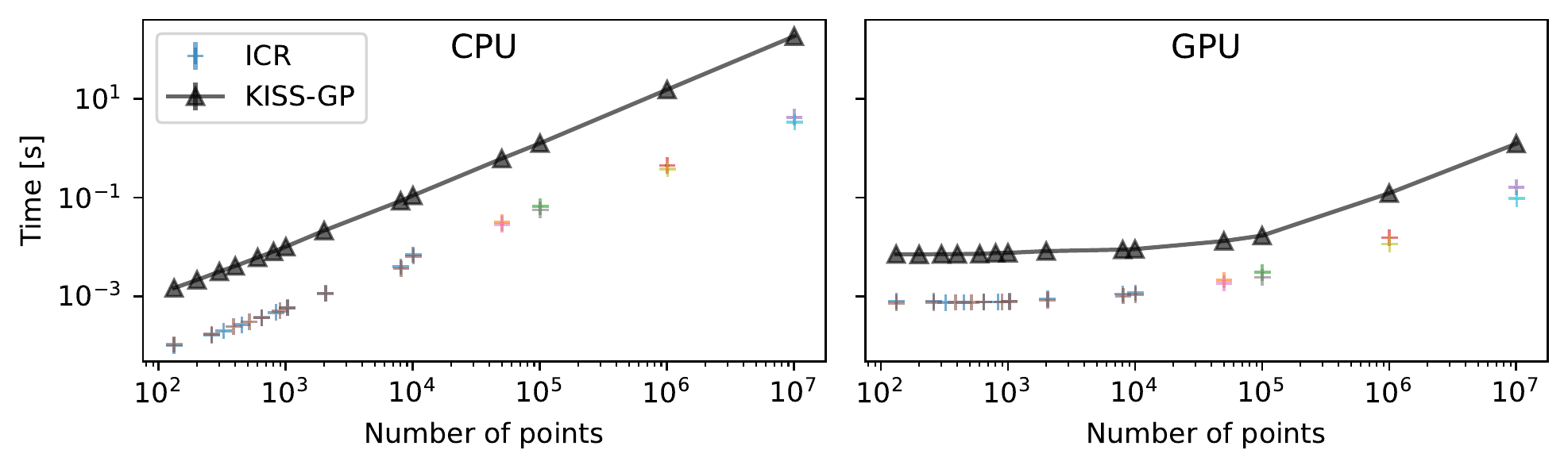}
	\caption{
	Performance benchmark on the CPU (left) and GPU (right) of KISS-GP versus ICR on double logarithmic axes.
	KISS-GP measurements are shown as triangles and ICR measurements as plus signs.
	The markers are placed at the median time it took the model to perform a forward pass.
	Minimum and maximum timings are shown as vertical bars but are fully absorbed by the marker size.
	Each parametrization of ICR is denoted by its own line style and color.
	\label{fig:refinement_chart_versus_SKI_KISS_benchmark}%
	}
\end{figure}

\autoref{fig:refinement_chart_versus_SKI_KISS_benchmark} depicts the execution time of one forward pass of the KISS-GP model versus ICR for a varying number of modeled points for the CPU and GPU on double logarithmic axes.
ICR is consistently about one order of magnitude faster than KISS-GP on both the CPU and GPU.
Its computational advantage changes minimally with the choice of parameters for ICR, which are denoted by different colors and line styles in the figure.

\section{Conclusion}
\label{sec:conclusion}

We introduce a new Iterative Charted Refinement (ICR) algorithm for evaluating GPs without any nested optimizations.
Taking a generative perspective on GPs, we reformulate the problem of applying the inverse and computing the log-determinant of the kernel matrix to applying the ``square-root'' of it.
We devise an algorithm that iteratively refines points on arbitrarily charted coordinates with $\mathcal{O}(N)$ computational complexity.
Its major advantages are its guaranteed positive semidefinite kernel matrix of full-rank and its computational speed.
For points whose spacings vary over two orders of magnitude, the representation of the kernel matrix in terms of the absolute pixel wise difference is comparable though slightly worse to KISS-GP's (in general, singular) kernel matrix representation.
In terms of computational speed, ICR consistently outperforms KISS-GP by about one order of magnitude on both the CPU and GPU.

In this paper we take first steps towards evaluating GPs via representation and application of the square-root of the kernel matrix without grid constraints.
Our method requires a user-provided coordinate chart to discretize the modeled locations at varying resolutions.
ICR is especially well suited for applications with highly unevenly spaced points and in which a full rank kernel matrix and computational speed is of upmost importance.
It has already been successfully applied to a GP with 122 billion degrees of freedom~\cite{Icecone:Leike2022}, and we envision many similar applications to follow.


\section*{References}

\printbibliography[heading=none]

@inproceedings{KernelInterpolation:Wilson2015,
	author = {Andrew Gordon Wilson and Hannes Nickisch},
	title = {Kernel Interpolation for Scalable Structured Gaussian Processes {(KISS-GP)}},
	year = {2015},
	% series = {32nd International Conference on Machine Learning, ICML 2015},
	publisher = {International Machine Learning Society (IMLS)},
	pages = {1775--1784},
	% editor = {Francis Bach and David Blei},
	booktitle = {32nd International Conference on Machine Learning, ICML 2015},
	% note = {32nd International Conference on Machine Learning, ICML 2015 ; Conference date: 06-07-2015 Through 11-07-2015},
	url = {http://proceedings.mlr.press/v37/wilson15.pdf},
}

@article{WhenGaussianProcessesMeetsBigData:Liu2019,
	title={When Gaussian Process Meets Big Data: A Review of Scalable GPs},
	author={Haitao Liu and Y. Ong and Xiaobo Shen and Jianfei Cai},
	journal={IEEE Transactions on Neural Networks and Learning Systems},
	year={2020},
	volume={31},
	pages={4405-4423},
	doi={10.1109/TNNLS.2019.2957109},
}

@inproceedings{ExactGaussianProcessesOnAMillionDataPoints:Wang2019,
	author = {Wang, Ke and Pleiss, Geoff and Gardner, Jacob and Tyree, Stephen and Weinberger, Kilian Q and Wilson, Andrew Gordon},
	booktitle = {Advances in Neural Information Processing Systems},
	% editor = {H. Wallach and H. Larochelle and A. Beygelzimer and F. d\textquotesingle Alch\'{e}-Buc and E. Fox and R. Garnett},
	pages = {},
	publisher = {Curran Associates, Inc.},
	title = {Exact Gaussian Processes on a Million Data Points},
	url = {https://proceedings.neurips.cc/paper/2019/file/01ce84968c6969bdd5d51c5eeaa3946a-Paper.pdf},
	volume = {32},
	year = {2019}
}

@inproceedings{FastKernelLearningForMultidimensionalPatternExtrapolation:Wilson2014,
	author = {Wilson, Andrew G and Gilboa, Elad and Nehorai, Arye and Cunningham, John P},
	booktitle = {Advances in Neural Information Processing Systems},
	% editor = {Z. Ghahramani and M. Welling and C. Cortes and N. Lawrence and K.Q. Weinberger},
	pages = {},
	publisher = {Curran Associates, Inc.},
	title = {Fast Kernel Learning for Multidimensional Pattern Extrapolation},
	url = {https://proceedings.neurips.cc/paper/2014/file/77369e37b2aa1404f416275183ab055f-Paper.pdf},
	volume = {27},
	year = {2014}
}

@inproceedings{GaussianProcessesForBigData:Hensman2013,
	author = {Hensman, James and Fusi, Nicol\`{o} and Lawrence, Neil D.},
	title = {Gaussian Processes for Big Data},
	year = {2013},
	publisher = {AUAI Press},
	% address = {Arlington, Virginia, USA},
	booktitle = {Proceedings of the Twenty-Ninth Conference on Uncertainty in Artificial Intelligence},
	pages = {282–290},
	% numpages = {9},
	% location = {Bellevue, WA},
	% series = {UAI'13}
	url = {https://proceedings.mlr.press/v37/wilson15.html},
}

@inproceedings{KernelInterpolationForScalableStructuredGaussianProcessesKISSGP:Wilson2015,
	author = {Wilson, Andrew Gordon and Nickisch, Hannes},
	title = {Kernel Interpolation for Scalable Structured Gaussian Processes (KISS-GP)},
	year = {2015},
	publisher = {JMLR.org},
	booktitle = {Proceedings of the 32nd International Conference on International Conference on Machine Learning - Volume 37},
	pages = {1775–1784},
	% numpages = {10},
	% location = {Lille, France},
	% series = {ICML'15}
}

@article{ThoughtsOnMassivelyScalableGaussianProcesses:Wilson2015,
	doi = {10.48550/ARXIV.1511.01870},
	title={Thoughts on Massively Scalable Gaussian Processes},
	author={Andrew Gordon Wilson and Christoph Dann and Hannes Nickisch},
	journal={arXiv},
	year={2015},
	volume={abs/1511.01870},
}

@inproceedings{ADVI:Warren2017,
	title = { Automatic Differentiation Variational Inference with Mixtures },
	author = {Morningstar, Warren and Vikram, Sharad and Ham, Cusuh and Gallagher, Andrew and Dillon, Joshua},
	booktitle = {Proceedings of The 24th International Conference on Artificial Intelligence and Statistics},
	pages = {3250--3258},
	year = {2021},
	% editor = {Banerjee, Arindam and Fukumizu, Kenji},
	volume = {130},
	% series = {Proceedings of Machine Learning Research},
	month = {04},
	publisher = {PMLR},
	url = {https://proceedings.mlr.press/v130/morningstar21b.html},
}

@article{MGVI:Knollmueller2019,
	doi = {10.48550/ARXIV.1901.11033},
	author = {Knollmüller, Jakob and Ensslin, Torsten A.},
	title = {Metric Gaussian Variational Inference},
	journal = {arXiv},
	year = {2019},
	volume = {abs/1901.11033},
}

@article{geoVI:Frank2021,
	doi = {10.3390/e23070853},
	year = 2021,
	month = {07},
	publisher = {{MDPI} {AG}},
	volume = {23},
	number = {7},
	pages = {853},
	author = {Philipp Frank and Reimar Leike and Torsten A. En{\ss}lin},
	title = {Geometric Variational Inference},
	journal = {Entropy},
}

@inproceedings{VariationalInferenceWithNormalizingFlows:Rezende2015,
	author = {Rezende, Danilo Jimenez and Mohamed, Shakir},
	title = {Variational Inference with Normalizing Flows},
	year = {2015},
	publisher = {JMLR.org},
	booktitle = {Proceedings of the 32nd International Conference on International Conference on Machine Learning - Volume 37},
	pages = {1530–1538},
	% numpages = {9},
	location = {Lille, France},
	series = {ICML'15},
	url = {http://proceedings.mlr.press/v37/rezende15.html},
}

@article{OptimalBeliefApproximation:Reimar2017,
	AUTHOR = {Leike, Reimar H. and Ensslin, Torsten A.},
	TITLE = {Optimal Belief Approximation},
	JOURNAL = {Entropy},
	VOLUME = {19},
	YEAR = {2017},
	NUMBER = {8},
	% ARTICLE-NUMBER = {402},
	ISSN = {1099-4300},
	DOI = {10.3390/e19080402},
}

@article{FoundationsOfInference:Knuth2012,
	author = {Knuth, Kevin H. and Skilling, John},
	title = {Foundations of Inference},
	doi = {10.3390/axioms1010038},
	year = {2012},
	month = {06},
	publisher = {{MDPI} {AG}},
}

@misc{Icecone:Leike2022,
	doi = {10.48550/ARXIV.2204.11715},
	author = {Leike, R. H. and Edenhofer, G. and Knollmüller, J. and Alig, C. and Frank, P. and Ensslin, T. A.},
	title = {The Galactic 3D large-scale dust distribution via Gaussian process regression on spherical coordinates},
	publisher = {arXiv},
	year = {2022},
}

@book{AnIntroductionToMultigridMethods:Wesseling2004,
	title={An Introduction to Multigrid Methods},
	author={Wesseling, P.},
	isbn={9781930217089},
	series={An Introduction to Multigrid Methods},
	url={https://books.google.de/books?id=RoRGAAAAYAAJ},
	year={2004},
	publisher={R.T. Edwards}
}

@book{ScientificComputing:Heath2018,
	title={Scientific Computing: An Introductory Survey, Revised Second Edition},
	author={Heath, Michael T},
	year={2018},
	publisher={SIAM},
	isbn = {1611975573},
	doi = {10.2514/1.J060261},
}

@article{OnTheConvergenceOfARelaxationMethodWithNaturalConstraintsOnTheEllipticOperator:Bakhvalov1966,
	title={On the convergence of a relaxation method with natural constraints on the elliptic operator},
	author={Bakhvalov, Nikolai Sergeevich},
	journal={USSR Computational Mathematics and Mathematical Physics},
	volume={6},
	number={5},
	pages={101--135},
	year={1966},
	publisher={Elsevier}
}

@inproceedings{GPyTorch:Gardner2018,
	author = {Gardner, Jacob and Pleiss, Geoff and Weinberger, Kilian Q and Bindel, David and Wilson, Andrew G},
	booktitle = {Advances in Neural Information Processing Systems},
	% editor = {S. Bengio and H. Wallach and H. Larochelle and K. Grauman and N. Cesa-Bianchi and R. Garnett},
	pages = {},
	publisher = {Curran Associates, Inc.},
	title = {GPyTorch: Blackbox Matrix-Matrix Gaussian Process Inference with GPU Acceleration},
	url = {https://proceedings.neurips.cc/paper/2018/file/27e8e17134dd7083b050476733207ea1-Paper.pdf},
	volume = {31},
	year = {2018}
}

@inproceedings{ConstantTimePredictiveDistributionsForGaussianProcesses:Pleiss2018,
	title = {Constant-Time Predictive Distributions for {G}aussian Processes},
	author = {Pleiss, Geoff and Gardner, Jacob and Weinberger, Kilian and Wilson, Andrew Gordon},
	booktitle = {Proceedings of the 35th International Conference on Machine Learning},
	pages = {4114--4123},
	year = {2018},
	% editor = {Dy, Jennifer and Krause, Andreas},
	volume = {80},
	series = {Proceedings of Machine Learning Research},
	month = {07},
	publisher = {PMLR},
	url = {https://proceedings.mlr.press/v80/pleiss18a.html},
}

@inproceedings{StochasticVariationalDeepKernelLearning:Wilson2016,
	author = {Wilson, Andrew G and Hu, Zhiting and Salakhutdinov, Russ R and Xing, Eric P},
	booktitle = {Advances in Neural Information Processing Systems},
	% editor = {D. Lee and M. Sugiyama and U. Luxburg and I. Guyon and R. Garnett},
	pages = {},
	publisher = {Curran Associates, Inc.},
	title = {Stochastic Variational Deep Kernel Learning},
	url = {https://proceedings.neurips.cc/paper/2016/file/bcc0d400288793e8bdcd7c19a8ac0c2b-Paper.pdf},
	volume = {29},
	year = {2016}
}

@inproceedings{SparseGaussianProcessesUsingPseudoInputs:Snelson2005,
	author = {Snelson, Edward and Ghahramani, Zoubin},
	booktitle = {Advances in Neural Information Processing Systems},
	editor = {Y. Weiss and B. Sch\"{o}lkopf and J. Platt},
	pages = {},
	publisher = {MIT Press},
	title = {Sparse Gaussian Processes using Pseudo-inputs},
	url = {https://proceedings.neurips.cc/paper/2005/file/4491777b1aa8b5b32c2e8666dbe1a495-Paper.pdf},
	volume = {18},
	year = {2005},
}

@inproceedings{FastMatrixSquareRootsWithApplicationsToGaussianProcessesandBayesianOptimization:Pleiss2020,
	author = {Pleiss, Geoff and Jankowiak, Martin and Eriksson, David and Damle, Anil and Gardner, Jacob},
	booktitle = {Advances in Neural Information Processing Systems},
	editor = {H. Larochelle and M. Ranzato and R. Hadsell and M.F. Balcan and H. Lin},
	pages = {22268--22281},
	publisher = {Curran Associates, Inc.},
	title = {Fast Matrix Square Roots with Applications to Gaussian Processes and Bayesian Optimization},
	url = {https://proceedings.neurips.cc/paper/2020/file/fcf55a303b71b84d326fb1d06e332a26-Paper.pdf},
	volume = {33},
	year = {2020},
}

@software{jax2018github,
	author = {James Bradbury and Roy Frostig and Peter Hawkins and Matthew James Johnson and Chris Leary and Dougal Maclaurin and George Necula and Adam Paszke and Jake Vander{P}las and Skye Wanderman-{M}ilne and Qiao Zhang},
	title = {{JAX}: composable transformations of {P}ython+{N}um{P}y programs},
	url = {http://github.com/google/jax},
	version = {0.3.8},
	year = {2022},
}

@software{nifty2022mpcdf,
	author = {Philipp Arras and Gordian Edenhofer and Philipp Frank and Andrija Kostic and Jakob Knollmüller and Jakob Roth and Lukas Platz and Matteo Guardiani and Martin Reinecke and Reimar Heinrich Leike and Simon Ding and Vincent Eberle},
	title = {{NIFTy}: Numerical Information Field Theory},
	url = {https://gitlab.mpcdf.mpg.de/ift/nifty},
	version = {8.0},
	year = {2022},
}

\end{document}